# Optimal Route Planning with Prioritized Task Scheduling for AUV Missions

S. MahmoudZadeh, D. Powers, K. Sammut, A. Lammas, A.M. Yazdani

Centre for Maritime Engineering, Control and Imaging

Flinders University, Adelaide, SA 5042, Australia

*Abstract*—**This paper presents a solution to Autonomous Underwater Vehicles (AUVs) large scale route planning and task assignment joint problem. Given a set of constraints (e.g., time) and a set of task priority values, the goal is to find the optimal route for underwater mission that maximizes the sum of the priorities and minimizes the total risk percentage while meeting the given constraints. Making use of the heuristic nature of genetic and swarm intelligence algorithms in solving NP-hard graph problems, Particle Swarm Optimization (PSO) and Genetic Algorithm (GA) are employed to find the optimum solution, where each individual in the population is a candidate solution (route). To evaluate the robustness of the proposed methods, the performance of the all PS and GA algorithms are examined and compared for a number of Monte Carlo runs. Simulation results suggest that the routes generated by both algorithms are feasible and reliable enough, and applicable for underwater motion planning. However, the GA-based route planner produces superior results comparing to the results obtained from the PSO based route planner.**

## I. Introduction

Autonomous Underwater Vehicles (AUVs) have been widely developed in the last few decades. AUVs have the potential of exploration in unknown undersea environments and today are the first choice to navigate autonomously and undertake various missions. AUVs' failure in underwater missions is not acceptable because maintenance is usually difficult and very expensive. Thus AUV should possess intelligent decision-making to carry out a given mission in a hazardous undersea environment before it runs out of time and battery, so mission timing is extremely critical to mission success.

AUVs are deployed to complete tasks such as water pollution/mineral monitoring, geological sampling, mosaicking the seafloor, underwater navigation, trajectory tracking and so on [1]. This makes time and task management challenging, considering mission type and time, number of task to complete, problem restrictions, time limitations versus changing environmental conditions and energy endurance. AUVs should carry out complex tasks in a pre-specified time interval. Hence, they have to effectively manage the available time for a series of deployment involving long missions. This management depends tightly on the optimality of the selected route between start and destination point. Thus, route planning for AUVs in a large scale environment is a significant issue in mission success.

In this paper, vehicles route planning means finding an optimal route for waypoint guidance of an AUV considering the problem specifications, where the edge (distance) between each pair of waypoints represents a specific task (with relevant parameters). So the efficiency of the generated route should be evaluated relative to satisfaction of the specified criteria for the problem [2]. The route planner operates reactively (online) during the mission, therefore time optimality is critical in this approach [3, 4]. The optimum route may have several alternatives and generally contains a sequence of waypoints.

Vehicle route planning is categorized as an NP-hard problems due to the combinatorial nature of this problem and topology complexity of operational network. Obtaining the optimal solutions for NP-hard problems is computationally challenging issue and difficult to solve in practice. Generally, proposed solutions for mission route planning approach can be categorized into three main groups: grid-based methods, graph based strategies, and artificial intelligence based techniques [5]. The grid-based strategies are inefficient in cases where the workspace is very large or complex because the large numbers of cells render such solutions intractable. On the other hand, topology-based (graph-based) methods, which are very popular, usually look for the shortest route between two points in a network (graph). The major drawback of these methods is that they are time consuming owing to redundant computations and makes them expensive in terms of time complexity [6]. Some of the popular graph search algorithms like A* [7, 8, 9, 10] or Dijkstra [11, 12] operate based on cell decomposition and determine the cell-based route from the start to the destination point. Another category of methods used for mission route planning is the artificial and computational intelligence (AI and CI) approaches. While various deterministic techniques have been developed over three last decades, evolution-based, heuristic and meta-heuristic methods still remain appropriate possibilities for real time applications with larger dimensionality. Genetic and evolutionary methods have been explored for route generation for unmanned aerial vehicles to minimize fuel consumption for the mission [13]. A niche genetic algorithm (INGA) improved real-time route planning of unmanned aerial vehicles [14]. Subsequently an offline pre-generative route planning strategy based on the non-dominated sorting genetic algorithm (NSGA-II) was proposed [15], but offline route generation strategies get in to serious difficulties when replanning is need due to problem arising during the mission. Another model based on various types of fuzzy arc lengths designed by [16] to compute the shortest route in a graph. Due to the high complexity of this method for larger problems, the GA used to find the shortest path relative membership function in the graph. Later on, a GA [17] and a hybrid Dijkstra GA based approach [18, 19] used to address the shortest path problems as combinatorial optimization problems. The results affirmed that Dijkstra consumes more time in finding optimum route comparing GA. Generally, evolutionary algorithms like GA [20, 21], PSO [22, 28] have low sensitivity to graph complexity, so search time increases linearly with the number of points.

Most of the reported route-finding strategies are single-objective, whilst in fact optimal route finding is mostly a multi-objective problem due to the existence of several cost factors such as route length, travel time, task priority and task specific metrics that to be simultaneously minimized or maximized. Unlike previous research on vehicle routing problems, which mostly look for the shortest possible path in a graph, this research aims to complete the maximum number of tasks for which time and distance are a function of the individual task. This problem require making maximum use of the available time but not exceeding it, rather than looking for a shortest path or accepting any feasible path. As AUV operates in an uncertain environment, there is a huge amount of uncertainty in the travel times that can have a devastating effect on mission plans. Proper time management of the vehicles routing operations is necessary ensure on-time mission completion and consequently the mission success.

The present research is about single vehicles operation, and explicitly assumes that it is not possible to cover all tasks in a single mission. Therefore, available tasks are prioritized in a way that selected edges (tasks) of the graph can take the AUV to the destination, which is a joint discrete and syndetic spaced problem at the same time. In this context, the proposed route planning problem can be modeled as a multi-objective optimization problem. It is thus necessary to address determination of a time optimum route between start and destination points in a large scale environment (i.e., *10 km² ×100m (depth)*), and carrying out maximum number of highest priority tasks (with small risk percentage). Generally, the task assignment (allocation) involves the decision making procedures under specified constraints and categorized as the complex combinatorial optimization problem [23]. For this purpose, the available mission time should be used as productively as possible, but the total travel time of the route should not exceed the overall mission available time.

Many deterministic algorithms and graph search methods have been introduced for solving the route planning or task assignment problems. The deterministic methods produce better quality solutions, however these algorithms are computationally complex. Therefore they are not appropriate approaches for real-time routing applications, specifically when the operating graph is topologically complex [24, 25]. In contrast, the meta-heuristic methods take less computation time and obtains optimal or near optimal solutions quickly. To cover objectives of this research two evolution-based approaches have been used to find the optimum route in the operating area with respect to problem objectives and constraints. The GA is one of the fastest optimization algorithms and it is well suited to graph searching problems due to its discrete nature. GA based approaches propose appropriate solution for complex graph routing problems in real-time applications.

This research takes the advantage of GA and PSO algorithms to solve vehicle planning problems according to defined objectives and specific restrictions. The main problem with the PSO implementation is proper coding of the particles as each particle in going to propose a valid route candidate. Due to the discrete nature of the search space, a particular problem arises using PSO, as it operates in a continuous space originally. However, the argument for using PSO is a strong one as it does appear to scale well with problem complexity, and can naturally encode the multi-objective nature of this problem. To solve the raised problem with PSO, in the considered case, this research contributes a priority based route generation approach on the underlying search space. The generated feasible routes have been encoded into particles based on priority and Adjacency matrixes (the detail discussed in section 4). These modifications increase the speed of the algorithm in finding optimum solution and prevent stucking in a local optima.

The organization of the paper is as follows. In section II, formulation is demonstrated. PSO and GA paradigm is briefly discussed in section III and IV. Section V describes the particle encoding mechanism and the overall process of PSO on carrying out the discussed problem. The discussion on simulation results are provided in Section VI. And, the section VII concludes the paper.

II. PROBLEM FORMULATION

The problem to be solved is ideally to find the optimum route covering the maximum number of highest priority tasks with smallest risk percentage in a time interval that battery's capacity allows. Optimality of the mission routes is subject to several constraints and objectives, and generally is a tradeoff or pareto problem. The planned route should be applicable and logically feasible, according to feasibility criteria's given in section V. In the initial study, it is assumed that the vehicle is moving with constant velocity in a 3D environment comprising several fixed waypoints. An underwater mission is commenced at a specified starting point and it is terminated when the AUV reaches to a predefined destination point. The vehicle should carry out the maximum number of tasks in available time and ensure it reaches to the destination before running out of time. Tasks assigned to edges of the graph in advance. Each task involves three parameters of priority, risk percentage and required completion time. When a route generated, the optimality of the produced route should be evaluated based on traveling time, number of tasks completed, and total quality of the solutions based on priority and risk percentage of each edge. The optimum selected route should contain the highest priority tasks with minimum risk percentage among all.

AUV starts its mission from point ($WP_1$) with initial position of ($x_1, y_1, z_1$) and should pass sufficient number of waypoints to reach on the destination ($WP_n$) at ($x_n, y_n, z_n$). Waypoints in the terrain are connected with an edge like $q_i$ from a set of $q=\{q_1,...,q_m\}$, where $m$ is the number of edges in the graph. Each edge of the network like $q_i$ is assigned with a specific task from a set of $Task=\{Task_1,...,Task_k\}$. Each task has a value like $\rho_i$ from a limited set of $\rho=\{\rho_1,...,\rho_k\}$ that represents its priority comparing other tasks, and completion time of $\delta_T$ regardless of required time for passing the relevant edge. Each task also has a risk percentage of $\xi_T$ regardless of terrain hazards and risks. All these information can be represented in a graph form for better understanding of the problem as depicted in *Fig.1*. The route can be represented as $R_i=(x_1,y_1,z_1,...,x_i,y_i,z_i,...,x_n,y_n,z_n)$, where ($x_i,y_i,z_i$) is the coordinate of any arbitrary waypoint in geographical frame.

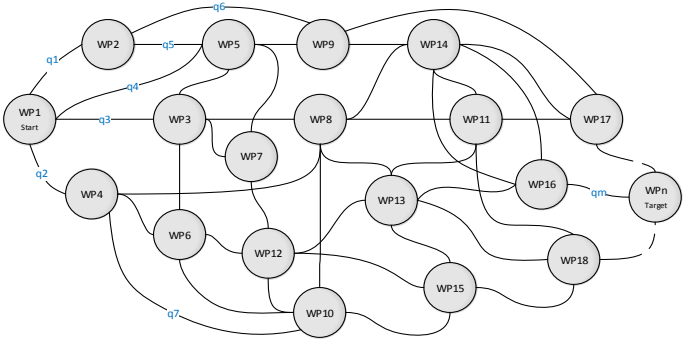

*Fig. 1.* A graph representation of operating area covered by waypoints

As previously discussed, the problem involves multiple objectives that should be satisfied during the optimization process. One approach in solving multi-objective problems is using multi-objective optimization algorithms. Another alternative is to transform a multi-objective optimization problem into a constrained single-objective problem. In this regard, the objective function is defined in a form of hybrid cost function comprising weighted functions that are required to be maximized or minimized. More detail about the cost function employed is expressed in section IV. In the preceding discussion, it is essential to describe the mathematical representation of the route planning problem for AUV in 3D environment. Therefore, it describes as follows:

$$q_{ij} : \begin{cases} d_{ij} \\ t_{ij} \end{cases} \qquad Task_{q_{ij}} : \begin{cases} \rho_{T_{ij}} \\ \delta_{T_{ij}} \\ \xi_{T_{ij}} \end{cases} \quad (1)$$

$$d_{ij} = \sqrt{(x_j - x_i)^2 + (y_j - y_i)^2 + (z_j - z_i)^2}$$

$$t_{ij} = \frac{d_{ij}}{V_{AUV}} + \delta_{Tij} \quad (2)$$

$$T_{Route} = \sum_{\substack{i=0 \\ j \neq i}}^{n} lq_{ij} t_{ij} = \sum_{\substack{i=0 \\ j \neq i}}^{n} lq_{ij} \left( \frac{d_{ij}}{V_{AUV}} + \delta_{Tij} \right), \quad l \in \{0,1\} \quad (3)$$

$$T_{travel} \leq T_{Route} < T_{available} \quad (4)$$

Then:

i. The total weight of route should be maximized:

$$W_R = \sum_{\substack{i=0 \\ j \neq i}}^{n} lq_{ij} \frac{\rho_{ij}}{\xi_{ij}} = \sum_{\substack{i=0 \\ j \neq i}}^{n} lq_{ij} w_{ij} \quad (5)$$

$$\max \left( \sum_{\substack{i=0 \\ j \neq i}}^{n} lq_{ij} w_{ij} \right), \quad l \in \{0,1\}$$

ii. The route travel time should approach total available time:

$$\min \left( |T_{Route} - T_{available}| \right) = \min \left( \left| \sum_{\substack{i=0 \\ j \neq i}}^{n} lq_{ij} \left( \frac{d_{ij}}{V_{AUV}} + \delta_{Tij} \right) - T_{available} \right| \right) \quad (6)$$

s.t:

total route travel time shouldn't exceed available mission time

$$\max(T_{Route}) < T_{available} \quad (7)$$

where $T_{Route}$ is the required time to pass the route, $T_{available}$ is the total mission time, $l$ is the selection variable, $t_{ij}$ is the required time to pass the distance $d_{ij}$ between two waypoint of $WP_i$ and $WP_j$ along with completion time of the task $\delta_{Tij}$ assigned to $q_{ij}$. $\rho_{Tij}$ and $\xi_{Tij}$ represent the priority and risk percentage of the task assigned to $q_{ij}$. $T_{travel}$ is the traveled time by AUV at each stage of mission.

III. OVERVIEW OF GENETIC ALGORITHM

Genetic Algorithm (GA) is a particular type of stochastic search algorithm represents problem solving technique based on biological evolution. GA has been extensively studied and widely used on many fields of engineering. GA provides alternative for traditional method that can be applied for nonlinear programming. GA search in a population space that each individual of this population is known as chromosome. Its process starts with randomly selecting a number of feasible solutions (chromosome) from initial population. A fitness function should be defined to evaluate each chromosome and quality of solution during the evolution process. Then, the set of best solution is selected from initial population using adaptive heuristic search nature of the GA. New population is generated from initial population using the GA operators like, selection, crossover and mutation. Chromosomes with the best fitness value are transferred to next generation and the rest will be eliminated. This progress continues until the chromosomes get the best fit solution to the given problem [26]. The average fitness of the population improves at each iteration, therefore after many iterations better solutions are revealed.

This route planning module deals with finding the optimal route through the operating graph using genetic algorithm. The input to this module is a group of feasible generated routes involving a sequence of nodes and all are common in starting and ending points and encoded as chromosomes. After primary population initialized, the algorithm starts its operating according to following pseudo code.

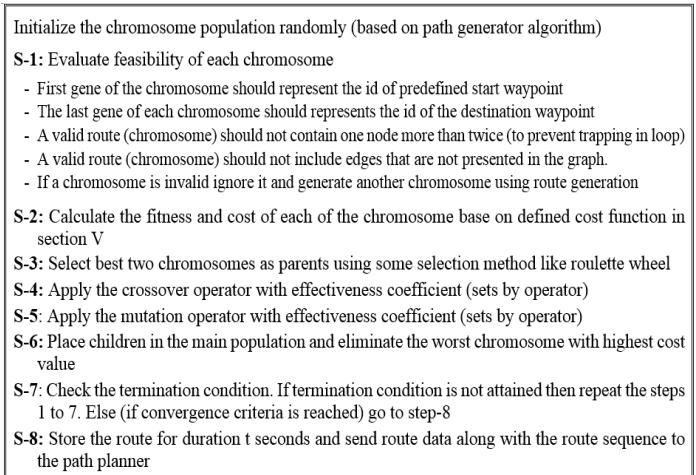

Fig. 2. GA optimal route generation pseudo code

## IV. OVERVIEW OF PARTICLE SWARM OPTIMIZATION

Particle swarm optimization is one of the fastest optimization methods for solving many complex problems widely used in several studies in past decades. The process in PSO is initialized with a population of particles. Each particle involves a position and velocity in the search space. The position and velocity of each particle gets updated in each iteration. Then, the performance of particles evaluated according to the fitness/cost functions. Each particle has memory for previous state values, its best position in its experience as $P_{best}$, and the global best position as $G_{best}$. In each iteration, the current state value of the particle is compared with $P_{best}$ and $G_{best}$. More detail about the algorithm can be found in related references [27]. Particle position and velocity get updated as follows (8) and (9):

$$\alpha_i(t+1) = \omega\alpha_i(t) + c_1 r_1 [P_{best_i}(t) - \lambda_i(t)] + c_2 r_2 [G_{best}(t) - \lambda_i(t)] \quad (8)$$

$$\lambda_i(t+1) = \lambda_i(t) + v_i(t+1) \quad (9)$$

where $c_1$ and $c_2$ are acceleration coefficients, $\lambda_i$ and $\alpha_i$ are particle position and velocity at iteration $t$. $P_{best-i}$ is the personal best position and $G_{best}$ is the global best position. $r_1$ and $r_2$ are two independent random numbers in $[0,1]$. $\omega$ exposes the inertia weight and balances the PSO algorithm between the local and global search. Due to discrete nature of current problems search space, a particular problem arises using PSO, as it operates in a continuous space originally. However, the argument for using PSO is a strong one as it does appear to scale well with problem complexity, and can naturally handle the multi-objective nature of this problem. To solve the raised problem with PSO, in the considered case, this research contributes a priority-adjunct based route generation approach on the underlying search space. The process of algorithm is presented in following flowchart:

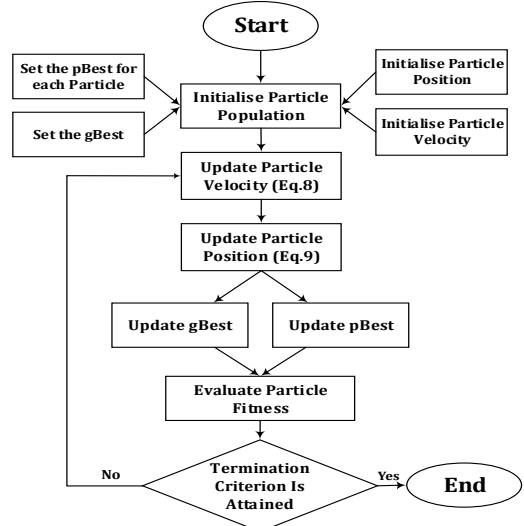

Fig. 3. The process of PSO algorithm

## V. IMPLEMENTATION

### A. Route Generation (Initialization Phase for both GA and PSO)

Suitable coding scheme for particle or chromosomes representation is the most critical step in formulating the problem in GA and PSO framework and it has direct impact on overall performance of the algorithm. The resultant solution from both GA and PSO should be feasible and valid according to following criteria's:

- A valid route should be commenced and ended with predefined start and target nodes.
- The generated route should not include edges that are not presented in the graph.
- The multiple appearance of the same node in a route makes it invalid because this issue implies wasting time repeating a task, which is a common problem with undirected graphs.
- Similarly, a route should not traverse an edge for more than once.
- The route travel time should not exceed the maximum range of AUVs' total available time.

Fig. 4. Route feasibility criteria

Therefore, a priority based strategy has been used in this research in order to generate feasible routes. For this purpose some guiding information of priority is added to each node at the initial phase. The priority vector initialized randomly. The sequence of nodes are selected based on their corresponding value in priority vector and adjacency matrix (adjacency matrix represents relations and edges in a graph). Then, to prevent generating infeasible routes some modifications have been applied as follows:

- Each node take positive or negative priority values in the specified range of [-100,100]. The selected node in a route sequence gets a large negative priority value that prevents repeated visits to a node. So that, the selected node will not be a candidate for future selection. This issue reduces the memory usage and time complexity for graphs with large number of nodes.

- Adjacency relations are used for adding nodes to a specific route, so nodes are added to the route sequence one by one according to priority vector and adjacency matrix.

- To satisfy the termination criteria of feasible route generation, if the route ends with a non-destination node and/or the length of the route exceeds the number of existed nodes in the graph, the last node of the sequence will be replaced by index of the destination node. This process keeps the generated route in feasible (valid) space.

*Fig.5* presents an example of the route generating process according to a sample adjacency matrix of a graph and a random priority array.

| | |
|---|---|
| Ad | Example of adjacency matrix for a graph with 18 nodes |
| n | Node index where $n=1$ is the start and $n=18$ is the destination point |
| $R^k_{Ui}$ | Partial route corresponding to the priority vector of a route including $k$, $k \in n$ nodes. |
| $U_i$ | Priority array (random no repeated vector in range of [-100,100]) |

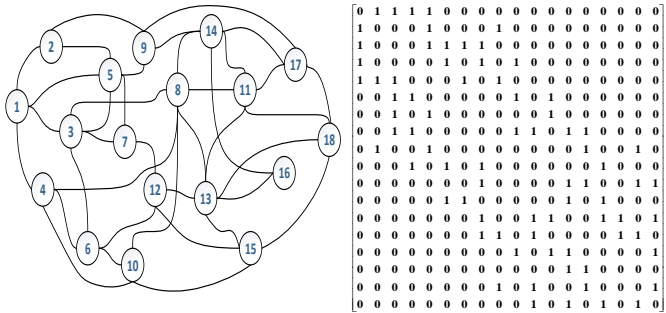

$U_i$ : {44, -38, 76, -78, 18, 47, 42, 61, 66, -69, -25, -93, 58, -15, 11, -43, 81, 97}

| Node index | 1 | 2 | 3 | 4 | 5 | 6 | 7 | 8 | 9 | 10 | 11 | 12 | 13 | 14 | 15 | 16 | 17 | 18 |
|---|---|---|---|---|---|---|---|---|---|---|---|---|---|---|---|---|---|---|
| $U_i \to R^1_{U_i} = \{1\}$ | -∞ | -38 | 76 | -78 | 18 | 47 | 42 | 61 | 66 | -69 | -25 | -93 | 58 | -15 | 11 | -43 | 81 | 97 |
| $R^2_{U_i} = \{1,3\}$ | -∞ | -38 | -∞ | -78 | 18 | 47 | 42 | 61 | 66 | -69 | -25 | -93 | 58 | -15 | 11 | -43 | 81 | 97 |
| $R^3_{U_i} = \{1,3,8\}$ | -∞ | -38 | -∞ | -78 | 18 | 47 | 42 | -∞ | 66 | -69 | -25 | -93 | 58 | -15 | 11 | -43 | 81 | 97 |
| $R^4_{U_i} = \{1,3,8,13\}$ | -∞ | -38 | -∞ | -78 | 18 | 47 | 42 | -∞ | 66 | -69 | -25 | -93 | -∞ | -15 | 11 | -43 | 81 | 97 |
| $R^4_{U_i} = \{1,3,8,13,18\}$ | -∞ | -38 | -∞ | -78 | 18 | 47 | 42 | -∞ | 66 | -69 | -25 | -93 | -∞ | -15 | 11 | -43 | 81 | -∞ |

Fig.5. A feasible route generation based on topological database (priority vector and adjacency matrix)

To find a route from start to destination node in a graph with 18 nodes based on a topological database, the first node will be selected and added to the sequence as the start position. Then from adjacency matrix the connected nodes to node-1 will be selected. In graph shown in *Fig.4*, this sequence is {*2,3,4,5*}. Then in this sequence the node with the highest priority (according to corresponding priority vector) will be selected and added to the route sequence as the next visited node. This procedure will be continued until a legitimate route is built (destination visited).

### B. Particle Swarm Optimization on Current Approach

Similar to GA, the PSO also get initialized based on route generation scheme proposed in section (A). Therefore, each particle assigned to a feasible route in the search space. Afterward the selected route should be evaluated according to defined cost function in t section D. The optimization process of the PSO-based algorithm for route planning is summarized as following pseudo-codes:

```
Initialize each particle by random velocity and position in following steps:
   – Assign a random priority array as Particle.Position
   – Initialize each particle with random velocity in range of [-20 20]
   – Extract a valid route from each particle according instructions mentioned in section(A)
   – If the generated route is infeasible, correct it according to mentioned modifications.
   – Evaluate generated route according to fitness/cost function.
   – If the route travel time exceeds the available mission time return a penalty value.
   – Calculate P_best and G_best for each particle.
S-1: Calculate new velocity for each particle according to Eq.8 ;
S-2: Calculate new position for each particle according to updated velocity in Eq.9;
S-3: Evaluate updated particle (route) according to defined cost function(given in section (D)).
S-4: If the route travel time exceeds the available mission time return a penalty value.
S-5: Compare P_best and current position for each particle and take the better one as P_best.
S-6: Check the termination condition. If termination condition is not attained then repeat the steps 1 to 6. Else (if convergence criteria is reached) go to step-7.
S-7: Store the route with relevant information
```

Fig. 6. PSO optimal route generation pseudo code

### C. Genetic Algorithm on Current Approach

#### 1) Chromosome Encoding

The chromosome in the proposed GA defined based on routes as sequence of nodes. The first and last gene of the chromosomes always corresponds to the start and destination node with respect to the topological information of the graph. The chromosomes take variable length, but limited to maximum number of nodes included in the graph, since it never required for a route to include nodes more than whole number of nodes in the graph.

#### 2) Selection

Selecting the parents for crossover and mutation operations is another step of the GA algorithm that plays an important role in improving the average quality of the population in the next generation. Several selection methods exist for this purpose such as roulette wheel selection, ranks election, elitist selection, scaling selection, tournament selection, etc. The roulette wheel selection has been conducted by current research, wherein the next generation is selected based on corresponding fitness or cost value, then the wheel divided into a number of slices and the chromosomes with the best cost take larger slice of the wheel.

*3) Crossover Operation*

Crossover is a GA operator that shuffles sub parts of two parent chromosomes and generate offspring that includes some part of both the parent chromosomes. Many types of crossover techniques have been suggested since now. Generally, they can be categorized in two main types of single point and multipoint crossover methods. In a single point crossover, only one crossing site existed, while in multipoint crossover, multiple sites of a pair of parent chromosomes are selected randomly to be shuffled. The single point crossover method is simple, but it has some drawbacks like formation of loop (cycles) when applied for routing problem. Therefore, to prevent such an issue it is required to use more advanced types of multipoint crossover methods like Order crossover (OX), Cycle crossover (CX), Partially Matched (PMX), Uniform crossover (UX) and so on [28]. Discussion over which crossover method is more appropriate to use still is an open area for research. Current research took advantages of uniform crossover, which uses a fixed mixing ratio among pair of parents and individual gens in the chromosomes are compared between two parents. The gens are swapped with a fixed probability that usually considers as 0.5. This method is extremely useful in problems with a very large search space in those where recombination order is important. After offspring generated, the new generation should be validated. Validation is carried out by checking the feasibility criteria defined for the routing problem then its fitness (or cost) is calculated. If the offspring does not correspond to a feasible route set, then it is eliminated from the next generation population.

*3) Mutation Operation*

Mutation is another operator that used in GA for generating the new population. This operator provides bit flipping, insertion, inversion, reciprocal exchange or others methods for generating new chromosomes from the parents [30]. Current research applies a combination of three inversion, insertion, and swapping types of mutation methods to generate the new population for GA. All these three methods preserve most adjacency information. In order to keep the new generation in feasible space, the mutation is applied on gens between but not included the first and last gens of the parent chromosomes that corresponds to start and destination point. After mutation operation is completed, the new offspring generated in this process have to be validated with the same procedure applied in crossover. Both of the mutation and crossover operations provide a search capability and enhance the rate of convergence.

*4) Termination Criteria*

The termination of the GA process can be defined according to completion of the maximum number of generations (Iterations), appearance of no change in population fitness after several iterations, and approaching to a stall generation. The most important step in finding an optimum route using GA is forming an appropriate and efficient cost function, so that the algorithm tends to compute the value of cost function for each route and provide a best fitted route with the maximum fitness and minimum cost value, since both are inversely proportional to each other. The cost evaluation is proposed in section D.

*D. Route Optimality Evaluation*

The cost function in this research is defined as a particular combination of the route traveling time, mission available time, task completion time, tasks priority and task risk percentage of each route. The cost function gets penalty *Vio* where the $T_{travel}$ for a particular route exceeds the available time for mission ($T_{available}$). The model is seeking an optimal solution in the sense of the best route according to given information. Thus, the total cost for the candidate route defined as (10):

$$Cost_{total} = \varphi_1 Cost_{Task} + \varphi_2 Cost_{Route} \tag{10}$$

in which $Cost_{Task}$ is the cost of task completion given in (14), $Cost_{Route}$ is the cost of generated route (12), $\varphi_1$ and $\varphi_2$ are two positive numbers that determine amount of participation of $Cost_{Task}$ and $Cost_{Route}$ on calculation of total cost.

$$Cost_{Route} = \left(|T_{Route} - T_{available}|\right) \times \left(1 + \gamma Viol\right) \tag{11}$$

$$Cost_{Route} = \left(\left|\sum_{\substack{i=0 \\ j \neq i}}^{n} lq_{ij}\left(\frac{d_{ij}}{V_{AUV}} + \delta_{Tij}\right) - T_{available}\right|\right) \times \left(1 + \gamma Viol\right) \tag{12}$$

$$Viol = \max\left(1 - \left(\frac{T_{available}}{T_{travel}}\right), 0\right) \tag{13}$$

where $\delta_{Tij}$ is the task completion time, $l$ is the selection parameter and takes value of 0 for unselected and 1 for selected edge. $T_{travel}$ is the time taken by the generated route and $T_{available}$ is the total mission available. $\gamma$ represents impact of *Violation* on total cost function.

$$Cost_{Task} = \left(\frac{\eta \sum \xi_{Route}}{\beta \sum \rho_{Route}}\right) \tag{14}$$

where $\xi_{Route}$ and $\rho_{Route}$ are the total risk percentage and priority of the tasks completed in generated route, $\eta$ and $\beta$ are coefficient that display the great importance of $\xi_{Route}$ and $\rho_{Route}$ in $Cost_{Task}$. Giving the appropriate value for engaged coefficients of factors in the cost function has a significant effect on the optimality of the generated route.

VI. SIMULATION RESULTS

The main purpose of the simulation experiments in this paper, is evaluating the performance of proposed PSO and GA based optimizers in generating real-time solution for vehicle routing and task assignment problem. A number of performance metrics have been investigated to evaluate the optimality of the proposed solutions. One of these metrics is the reliability percentage of the route including the chance of the mission success, which is combination of route violation value (whether it takes more time than entire available time) and validity of the generated route (bases on feasibility criteria's, *Fig.4*). Other metrics involve the number of completed tasks, total weight, total cost, and the time constraint satisfaction of the generated route with respect to the complexity of the graph presented in Table 1 and 2. The mission available time is set on 7 hours.

TABLE I. GRAPH COMPLEXITY AND GENERATED ROUTES (SOLUTIONS)

| Graph | Node | Edge | Solution | | Route |
|---|---|---|---|---|---|
| G1 | 50 | 1197 | 1 | PSO | [**1**,17,36,8,42,4,29,41,48,10,18,45,14,12,23,**50**] |
| | | | | GA | [**1**,35,46,42,48,407,10,18,36,23,13,15,28,33,45,29,12,**50**] |
| G2 | 100 | 4886 | 2 | PSO | [**1**,2,91,26,84,89,55,69,52,56,72,50,70,80,62,4093,3,**100**] |
| | | | | GA | [**1**,2,4,70,33,81,62,7,45,90,11,78,95,54,96,79,59,34,6,46,8 5,19,**100**] |

TABLE II. STATISTICAL ANALYZING OF THE ROUTE EVALUATION WITH PERFORMANCE METRICS

| Performance metrics | | Solution1 (50 Nodes) | | Solution 2 (100 Nodes) | |
|---|---|---|---|---|---|
| | | PSO | GA | PSO | GA |
| CPU Run Time(sec) | | 8.4 | 4.5 | 18.5 | 12.53 |
| Best Cost | | 0.033 | 0.023 | 0.036 | 0.0193 |
| Total Available Time(sec) | | 25200 (7h) | 25200 (7h) | 25200 (7h) | 25200 (7h) |
| Route Travel Time(sec) | | 23166 | 24176 | 25232 | 22605 |
| Total Distance | | 56515 | 61294 | 678532 | 61027 |
| Total Weight | | 38 | 52 | 49 | 58 |
| N-Tasks | | 16 | 19 | 18 | 23 |
| Reliability | Violation | 0.00 | 0.00 | 0.0043 | 0.00 |
| | Feasibility | Yes | Yes | Slightly late | Yes |

The PSO and GA configured with the same initial conditions and their performance are tested on graphs with same complexities including two cases with 50 nodes and 100 nodes. PSO and GA both proposed desirable optimal route with a quick computation time regardless of graph complexity. From simulation results in Table 1,2 and *Fig.7* it is noted that in all cases route travelling time is smaller than total available time (except PSO for 100 nodes which a slight violation exists) that confirms feasibility of the produced route. The provided results also confirm that the utilized methods are able to undertake the highest number of task and maximize the use of the available time (as $T_{travel}$ approaches $T_{available}$). Indeed it is noteworthy that the performance of both algorithms is relatively independent of both size and complexity of the graph, as this is a challenging problem for other algorithms. Hence, the algorithm is suitable to produce optimal solutions quickly for real-time applications and dynamic re-planning encountering environment dynamicity. Referring to *Fig.7* and Table 2 it is evident that both of presented methods can produce an optimum route considering performance metrics, however it is obvious that GA acts more efficiently in terms of computation time, minimizing cost value, total collected weights, and it covers more number of task. This cost presented in *Fig.7*, is produced by 250 iterations and same initial conditions for both PSO and GA on a graph with 100 nodes.

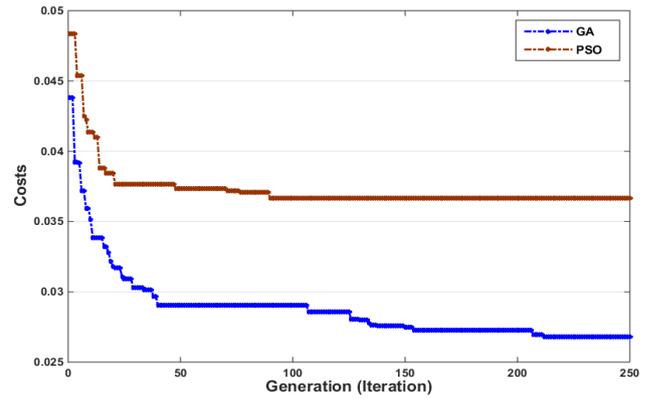

Fig. 7. Variation of cost for both PSO and GA in 100 iterations

To evaluate the robustness and reliability of the employed algorithms, 100 execution runs are performed in a Monte Carlo simulation based on total travel time and total obtained weight that is presented by *Fig.8*.

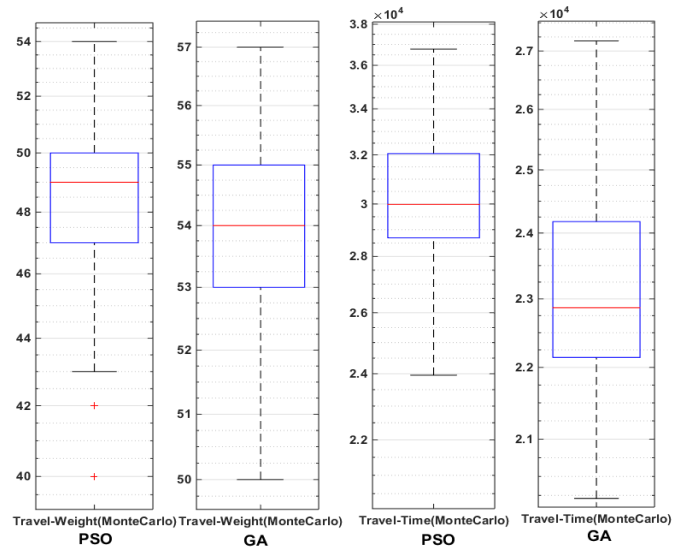

Fig. 8. Comparison between the performance of GA and PSO in terms of mission's metrics based on Monte Carlo simulation

The number of graph nodes is fixed on 20 waypoints for all Monte Carlo runs, but the topology of the graph was changed randomly based on a Gaussian distribution on the problem search space. The time threshold ($T_{available}$) also fixed on $3.06 \times 10^4 (sec)$. *Fig.8* demonstrates the functionality of GA and PSO in dealing with problem's space deformation and quantitative measurement of two significant mission's metrics, travel weight and time, which are directly associated with the number of successful task completion. As indicated in the graph, GA has superior performance and shows more consistency in its distribution. However, both algorithms reveal robust behavior to the variations and meet the specified constraint.

## VII. COSSNCLUSION

Global route planning along with task priority assignment are two important issues considered in mission time management and have great impact on mission success. The vehicle should generate an optimal route involving an appropriate number of waypoints, where the edge (distance) between each pair of waypoints represents a specific task including related parameters. This research investigated performance of a particle swarm optimization and genetic algorithm in providing time optimal routes while carrying out the mission goals under specific constraints. To fulfil the objectives of this research toward solving the stated problems, the solution was presented in several steps to enables the vehicle to autonomously find an optimal route through the operation network, carry out the maximum number of highest priority tasks, and reach to destination on time. Novel modification has been applied to route generating flow and route encoding distribution to prevent generating infeasible routes by reducing the possibility of loop-formation and speed up the entire process. Finally the system has simulated on different graphs of varying complexity.

The simulation results demonstrate that this new approach along with proposed algorithms could generate an optimal route in a very competitive CPU time. Indeed producing a real-time near optimal solution is more valuable than an optimal solution that takes too long. The performance of the solutions obtained by PSO method GA has been compared using the same configurations reinforcing that the proposed GA algorithm exhibits more desirable route optimality in a very competitive time. It is inferable from the result, the presented algorithms are not sensitive to the size of graph and they are able to produce optimum route in real-time applications. Future work will focus on development of a more efficient hybrid framework including global route planning and local path planning that dynamically takes into account the variable environment condition and different scenarios.